\documentclass[letterpaper]{article}
\usepackage[margin=0.75in]{geometry}
\usepackage[hyphens]{url}
\usepackage{graphicx}
\usepackage[numbers]{natbib}
\usepackage{caption}
\usepackage{booktabs}
\usepackage{amsmath,amssymb}
\usepackage{tabularx,array}
\usepackage{times}
\usepackage{newtxmath}
\newcommand{\supplementnotation}{%
\par\vskip 0.2em
\begingroup
\centering
  \footnotesize
  \setlength{\tabcolsep}{1.4pt}
  \renewcommand{\arraystretch}{0.82}
  \begin{tabular}{>{\raggedright\arraybackslash}p{0.12\textwidth}>{\raggedright\arraybackslash}p{0.35\textwidth}>{\raggedright\arraybackslash}p{0.12\textwidth}>{\raggedright\arraybackslash}p{0.35\textwidth}}
\toprule
\textbf{Symbol} & \textbf{Meaning} & \textbf{Symbol} & \textbf{Meaning} \\
\midrule
  \(x\) & Query or test case. & \(\mathcal A,K\) & Finite action set and number of actions.\\
  \(a,b\) & Candidate actions in an audited pair. & \(M_j,M\) & Source memory \(j\) and generic fitted memory.\\
  \(M_1\oplus M_2\) & Pooled multiset union of two memories. & \(\mathcal X_{\mathrm{test}}\) & Test-query set.\\
  \(a\succeq_x^M b\) & Readout preference relation. & \(\phi(x),\Phi_j\) & Query representation and source representation matrix.\\
  \(Q^M(x,a),\widehat v_a^M\) & Readout score and fitted action coefficient. & \(r_{j,a},d_j(a,b)\) & Source return vector and return-difference vector.\\
  \(y_j\) & Source evidence vector \(\Phi_j^\top d_j(a,b)\). & \(G_j,\widetilde G_j,G_0\) & Source raw, normalized, and reference Gram matrices.\\
  \(\delta_{\mathrm{norm}}\) & Scale-free normalized Gram mismatch. & \(G,y\) & Generic Gram matrix and pair-evidence vector in ridge perturbation.\\
  \(\Delta_j,\Delta_{12}\) & Source and pooled preference margins. & \(m_0,m_{\lambda_{\mathrm{ridge}}}\) & OLS and ridge-stabilized margins.\\
  \(\lambda_{\mathrm{ridge}},\mu\) & Ridge stabilization and \(\lambda_{\min}(G)\). & \(e=(x,\{a,b\})\) & Pairwise audit record.\\
\(\tau,\operatorname{dir}_\tau\) & Source-margin threshold and thresholded direction. & \(E_\tau,s_e\) & Eligible-pair set and shared strict direction.\\
\(\epsilon_{\mathrm{pair}},\epsilon_{\mathrm{tie}}\) & Pairwise and decision-tie tolerances. & \(\widehat{\mathrm{A2Cons}}_\tau\) & Empirical strict A2-consistency rate.\\
\(a_j,a_{12},a_s\) & Source, pooled, and source-consensus actions. & \(C,D\) & Source-consensus and decision-change sets.\\
\(u(x,a),\ell_u(x)\) & Task utility and signed utility loss. & \(h_D,h_{\mathrm{all}},h_C\) & Changed-decision, all-query, and consensus harmful rates.\\
\(\kappa_j,\kappa_\Sigma\) & Source Gram scale and \(\kappa_1+\kappa_2\). & \(\kappa_{\min},\kappa_{\max}\) & Minimum and maximum source Gram scales.\\
\(E_j,\epsilon,\epsilon_\delta\) & Gram perturbation and perturbation levels. & \(\lambda_0\) & Minimum eigenvalue of the reference Gram matrix \(G_0\).\\
\(G_\Sigma,H_j\) & Pooled proof matrix \(G_1+G_2\) and rescaled source matrix \((\kappa_\Sigma/\kappa_j)G_j\). & \(\|\cdot\|_F,\|\cdot\|_2\) & Frobenius and Euclidean/spectral norms.\\
\(B,\Phi_B,I_H\) & Mini-batch, batch representation matrix, and identity. & \(\Psi_B,\Gamma_B\) & Raw and normalized batch Gram matrices.\\
\(\Gamma_{\mathrm{low}}\) & Frozen lower-stage normalized Gram direction. & \(L_{\mathrm{task}},R,\lambda_{\mathrm{reg}}\) & Task loss, representation regularizer, and strength.\\
\(R_{\mathrm{white}},R_{\mathrm{direct}}\) & Whitening and direct-alignment penalties. & \(N,\alpha,H\) & Controlled memory size, anisotropy strength, and dimension.\\
\(c\) & Dataset class or target label for proxy returns. & \(t,i,h\) & Ordered load index, row/day index, and half-hour slot.\\
\(L_{i,h},\bar L_{i,h}\) & Observed and clipped normalized load. & \(\rho_{i,h},z_{i,h},m_{i,h}\) & Load ramp, base-price component, and market price.\\
\(\xi_{i,h},\eta_{i,h}\) & Gaussian price noise and rare spike component. & \(b_a,r(a,m)\) & Bid level and bidding-proxy return.\\
Acc. & Fraction of test queries for which the pooled action equals the task-defined oracle action. & & \\
\(|D|/|\mathcal X_{\mathrm{test}}|\) & All-query decision-change rate. & & \\
\bottomrule
\end{tabular}
\par\vskip 2pt
{\small Table A1: Main notation for the fixed-representation readout, strict composition audit, Gram-stability analysis, representation objectives, and reported experiments. The consolidated table separates source-specific, pooled, pairwise, query-level, and protocol quantities within one consistent vocabulary. This notation closes the formal-to-empirical loop and makes the paper's claims easier to verify.\par}
\endgroup}

\begin{document}
\begin{center}
{\LARGE More Data, Worse Decisions? Preference Reversals in Neural Networks under Gram Incompatibility\par}
\vspace{0.8em}
Manli Yan, Yuanzheng Li, Yong Zhao (Corresponding author), Hongbo Guo, Shoudong Han\\
Huazhong University of Science and Technology\\
Wuhan, China\\
\texttt{d202481469@hust.edu.cn, Yuanzheng\_Li@hust.edu.cn}\\
\texttt{zhiwei98530@hust.edu.cn, hongboguo@hust.edu.cn}\\
\texttt{shoudonghan@hust.edu.cn}\par
\vspace{0.8em}
Preprint\par
\end{center}
\vspace{0.6em}
\noindent Abstract\par\smallskip
Neural networks increasingly combine data across populations, time periods, and operating conditions to improve generalization. This raises a reliability question: whether a model refitted on pooled data preserves an action ordering supported by both sources. Case-Based Decision Theory (CBDT) formalizes this requirement through its composition axiom, which requires source-supported preferences to survive their union. We study when this property holds for fixed-representation neural networks with ordinary least squares (OLS) output heads. First, we show that pooled refitting recomputes the inverse-Gram geometry used to weight source evidence, which can reverse shared preferences, and derive exact and approximate preservation conditions. Next, we introduce a scale-invariant Gram mismatch measure for prioritizing candidate pools and geometry-oriented regularization for shaping source geometry during training. Finally, we develop a three-stage audit that traces strict pairwise reversals through decision changes to task-defined utility loss. Experiments spanning a load-based bidding proxy and medical and financial decision proxies reveal stable and reversal-prone pooling regimes: the load audit identifies a measurable nonzero class of source-consensus-relative harmful decisions under the proxy utility, while cross-domain audits show that comparable mismatch can correspond to sharply different preservation rates. Geometry-oriented objectives occupy distinct descriptive accuracy--consistency--geometry--harm operating points. Together, the framework makes compositional reliability measurable and operational through screening, analytic certification, geometry-oriented training, and decision-consequence auditing.\par
\vspace{0.8em}
\noindent Keywords: neural decision reliability; heterogeneous data pooling; compositional consistency; preference reversal; Gram incompatibility; case-based decision theory.
\vspace{1em}
\clearpage
\twocolumn

\section{1 Introduction}

Neural networks increasingly make decisions using data drawn from different populations, time periods, and operating conditions. Pooling such data broadens the evidence available to the model and can support better decisions. For this benefit to be reliable, however, pooling should not overturn agreements already present across sources. If both sources rank one action above another, a model refitted on their union should preserve that ordering. When a network selects among candidate actions, the choice depends on their relative scores; action ordering is therefore the natural object for evaluating whether multi-source integration improves rather than distorts neural network decisions.

Aggregate accuracy does not test this requirement: it measures predictive performance, not whether the relative ordering of actions remains consistent with source agreement. A pooled refit may therefore reverse an agreed ranking without that failure being reflected in aggregate accuracy. The consequences unfold in three stages. A pairwise reversal changes the decision only when it involves the source-consensus maximizer and moves a different action above it; even a changed decision may not reduce task-defined utility. Evaluation should therefore separate pairwise reversal, decision change, and utility loss.

Case-Based Decision Theory (CBDT) formalizes this requirement through its composition axiom: an ordering supported by both source memories should survive their union \cite{gilboa1995case,gilboa2012case}. Under fixed representations with ordinary least squares (OLS) output heads, action scores admit an exact case-weighted return form \cite{yan2026neuralcases}, making composition algebraically testable. Pooled refitting, however, recomputes the inverse Gram geometry used to weight source evidence and can reverse agreement. This leads to a natural progression: explain why reversals occur, prioritize candidate pools for audit, shape source geometry during training, and determine whether observed reversals reach decision consequences.

Accordingly, we make three contributions:

\begin{itemize}
\item
  Within CBDT, we formalize pooled-refit reliability as compositional preference preservation for fixed-representation neural networks with OLS output heads, show how inverse-Gram recomposition can reverse an ordering supported by both sources, and derive exact and approximate preservation conditions.
\item
  We introduce a scale-invariant Gram mismatch measure for screening candidate pools before refitting and geometry-oriented regularizers for shaping source geometry during training, enabling systematic comparison of accuracy--consistency--geometry--harm operating points.
\item
  We develop a three-stage audit that traces strict pairwise reversal through decision change to task-defined utility loss, evaluate the full consequence chain on a load-based bidding proxy, and use medical and financial proxies to characterize the scope and limitations of mismatch-based screening.
\end{itemize}

\section{2 Related Work}

Related work on neural decisions under data pooling falls into four linked areas: case support, heterogeneous-source learning, aggregation consistency, and representation alignment. We review them in this order.

Case-support research examines which training examples underlie a fitted neural prediction: case- and prototype-based methods use representative cases or learned prototypes, whereas influence-function and data-valuation methods trace predictions or losses to influential examples \cite{chen2019this,wolf2024keep,koh2017understanding,yeh2018representer,pruthi2020estimating}. These methods make support within a fitted model traceable. When neural networks are trained on data from different sources to improve generalization, however, the question shifts to whether source-specific models support the same action ordering and whether pooled refitting preserves that agreement.

Heterogeneous-source learning examines whether models transfer and retain performance across environments. Continual learning, domain adaptation, and distribution-shift studies evaluate transfer, robustness, forgetting, and degradation across sources \cite{kirkpatrick2017overcoming,he2023domain,koh2021wilds}. These evaluations characterize cross-source performance, but not whether pooled refitting preserves an ordering jointly supported by source-specific models.

Aggregation-consistency research examines what should survive combination and how combination can disrupt learned behavior. Preference-aggregation studies formulate consistency axioms \cite{hornischer2025learning,ge2024axioms}, while model-merging studies analyze interference and parameter-space compatibility \cite{cheng2025whoever,wei2025representation,chen2025pareto,wang2026mloss}. These perspectives are complementary but remain weakly connected: axiomatic work does not explain the geometry of pooled refitting, whereas parameter-space analyses do not directly address source-data aggregation under a shared representation.

Representation-alignment research examines whether source statistics can be made compatible. Normalization and covariance-alignment methods modify activation statistics or centered covariances to improve optimization, transfer, or stability \cite{ioffe2015batch,ba2016layer,huang2018decorrelated,sun2016deep}. Such methods can reduce distributional mismatch, but centered statistics do not fully characterize the raw, uncentered geometry relevant to refitted OLS heads or define a decision-level preservation criterion. These lines of work do not directly characterize whether source-supported action orderings survive pooled refitting under a shared representation, motivating the formalization that follows.

\section{3 Problem Formulation and Decision Audit}

This section establishes the formal basis for compositional reliability in fixed-representation neural networks with OLS output heads. We first express source-specific and pooled action orderings as preference margins, then turn CBDT's composition axiom into an empirical preservation rate, and finally trace how pairwise reversals propagate to decisions and task-defined utility.

\subsection{3.1 Fixed-Representation Neural Network with an OLS Output Head}

Each source dataset is represented as a memory \(M\), a finite multiset of cases in which each case contains a fixed neural representation and an associated return vector over the candidate actions; \(M_1\oplus M_2\) denotes multiset addition. The analysis is conditional on a trained representation: once representation training is complete, \(\phi\) is frozen while the source-specific and pooled readouts are fitted. Section 5 varies how this shared representation is trained before freezing; for any one reported configuration, all source-specific and pooled readouts use the same trained frozen representation. For each action \(a\), let \(\hat v_a^M\) be the OLS coefficient vector fitted on memory \(M\). Given the fixed representation \(\phi(x)\) of query \(x\), the fitted head assigns an action score and induces a preference relation:

\begin{equation}
\begin{aligned}
Q^M(x,a)&=\phi(x)^\top\widehat v_a^M,\\
a\succcurlyeq_x^M b&\Longleftrightarrow Q^M(x,a)\geq Q^M(x,b).
\end{aligned}
\tag{1}\label{eq:b-01}
\end{equation}

Thus, \(a\succeq_x^M b\) means that the readout fitted on memory \(M\) ranks action \(a\) at least as high as action \(b\) for query \(x\). We write the finite action set as \(\mathcal A=\{0,\ldots,K-1\}\). Any intercept is represented by augmenting the frozen representation with a constant coordinate, so the readout remains linear in \(\phi(x)\). A compact summary of the notation used throughout the paper is provided in Appendix A.

For source \(j\in\{1,2\}\), let \(\Phi_j\) stack its fixed representations row-wise, and let \(r_{j,a}\) collect the corresponding returns for action \(a\). The matrix \(G_j=\Phi_j^\top\Phi_j\) is the raw, uncentered Gram matrix of source \(j\), while \(d_j(a,b)=r_{j,a}-r_{j,b}\) collects the return differences between actions \(a\) and \(b\). Define the source preference margin as the score difference \(Q^{M_j}(x,a)-Q^{M_j}(x,b)\). When \(\Phi_j\) has full column rank, this margin is

\begin{equation}
\Delta_j^{a,b}(x)=\phi(x)^\top G_j^{-1}\Phi_j^\top d_j(a,b).
\tag{2}\label{eq:b-05}
\end{equation}

A positive margin means that source \(j\) supports \(a\) over \(b\); a negative margin supports the reverse ordering, and a zero margin denotes a tie.

Pooling \(M_1\) and \(M_2\) stacks their cases, so the Gram matrix becomes \(G_1+G_2\) and the two return-evidence vectors add. The resulting pooled preference margin is

\begin{equation}
\begin{aligned}
\Delta_{12}^{a,b}(x)
&=\phi(x)^\top(G_1+G_2)^{-1}\\
&\quad\cdot\left[\Phi_1^\top d_1(a,b)+\Phi_2^\top d_2(a,b)\right].
\end{aligned}
\tag{3}\label{eq:b-06}
\end{equation}

Here, \(\Phi_j^\top d_j(a,b)\) is the return-evidence vector contributed by source \(j\) for the action pair. Eqs. (2) and (3) isolate the composition mechanism: the source margins weight their respective evidence through \(G_j^{-1}\), whereas the pooled margin weights the combined evidence through the recomputed geometry \((G_1+G_2)^{-1}\).

Eqs. (2) and (3) give the exact full-rank OLS formulation. For an OLS margin \(m_0=\phi(x)^\top G^{-1}y\) and its ridge-stabilized counterpart \(m_{\lambda_{\mathrm{ridge}}}=\phi(x)^\top(G+\lambda_{\mathrm{ridge}} I)^{-1}y\), let \(\mu=\lambda_{\min}(G)>0\). The resolvent bound gives \(|m_{\lambda_{\mathrm{ridge}}}-m_0|\le \lambda_{\mathrm{ridge}}\|\phi(x)\|_2\|y\|_2/[\mu(\mu+\lambda_{\mathrm{ridge}})]\). Thus, ridge stabilization preserves the ordering whenever the corresponding OLS margin exceeds this perturbation in magnitude. The derivation is provided in Appendix B.6.

\subsection{3.2 Empirical Preference Preservation}

CBDT's composition axiom A2 turns source agreement into a preservation requirement. For a query \(x\) and actions \(a,b\), an ordering supported by both source memories should remain valid after their union:

\begin{equation}
a\succcurlyeq_x^{M_1}b\ \text{and}\ a\succcurlyeq_x^{M_2}b
\Longrightarrow a\succcurlyeq_x^{M_1\oplus M_2}b.
\tag{4}\label{eq:b-a2}
\end{equation}

A2 also preserves strictness: if either source preference is strict, the pooled preference is strict.

To evaluate A2 on finite data, let \(e=(x,\{a,b\})\) denote an unordered query-action pair. For \(e=(x,\{a,b\})\), write \(\Delta_j(e)=\Delta_j^{a,b}(x)\) and \(\Delta_{12}(e)=\Delta_{12}^{a,b}(x)\). For a margin threshold \(\tau\), let
\[
\operatorname{dir}_{\tau}(z)
=
\begin{cases}
1, & z>\tau,\\
-1, & z<-\tau,\\
0, & |z|\le\tau.
\end{cases}
\]
Set \(s_e=\operatorname{dir}_{\tau}(\Delta_1(e))\), and define
\[
E_{\tau}
=
\{e:s_e=\operatorname{dir}_{\tau}(\Delta_2(e))\ne0\}.
\]
All canonical audits use source threshold \(\tau=0\) and numerical tolerance \(\epsilon_{\mathrm{pair}}=10^{-6}\), separately record strict reversals and pooled ties, and compute the normalized Gram mismatch in Eq.~(8). The canonical audit focuses on the strict--strict subset of the A2 antecedent: both source margins must exceed the source threshold in the same direction. This choice avoids treating finite-precision source ties as stable support and yields a reproducible strict composition diagnostic. It does not exhaustively evaluate the weak--strict cases allowed by the full axiomatic statement in Eq.~(4). The empirical strict A2 consistency is
\begin{equation}
\widehat{\mathrm{A2Cons}}_{\tau}
=
\frac{1}{|E_{\tau}|}
\sum_{e\in E_{\tau}}
\mathbf{1}
\{s_e\Delta_{12}(e)>\epsilon_{\mathrm{pair}}\}.
\tag{5}\label{eq:b-07}
\end{equation}

Each eligible pair has a strict and shared source direction. The pooled readout preserves A2 only when its oriented margin remains strict beyond the numerical tolerance. A strict directional reversal satisfies \(s_e\Delta_{12}(e)<-\epsilon_{\mathrm{pair}}\), whereas \(|s_e\Delta_{12}(e)|\le\epsilon_{\mathrm{pair}}\) is recorded as a pooled-tie strictness failure. Thus, \(1-\widehat{\mathrm{A2Cons}}_\tau\) is the empirical A2-failure rate and coincides with the strict-reversal rate whenever the pooled-tie count is zero, as in the reported load and cross-domain audits; the isolated controlled-study ties are reported separately in Appendix D.

\subsection{3.3 From Pairwise Reversal to Decision Consequences}

The decision audit follows three stages across two statistical units. Pairwise preservation is evaluated over eligible query-action pairs, whereas decision changes and utility outcomes are evaluated over queries for which the two source models select the same action. For any action-score vector \(q\), decision ties use \(\epsilon_{\mathrm{tie}}=10^{-6}\) and the deterministic rule

\begin{equation}
\operatorname*{argmax}_{\epsilon_{\mathrm{tie}}}q
=\min\{a:q_a\geq\max_b q_b-\epsilon_{\mathrm{tie}}\}.
\tag{6}\label{eq:b-tie-rule}
\end{equation}

Let \(a_j(x)=\operatorname*{argmax}_{\epsilon_{\mathrm{tie}}}Q^{M_j}(x,\cdot)\) denote the action selected by source \(j\), and let \(a_{12}(x)=\operatorname*{argmax}_{\epsilon_{\mathrm{tie}}}Q^{M_1\oplus M_2}(x,\cdot)\) denote the pooled action. Define the source-consensus set \(C=\{x:a_1(x)=a_2(x)\}\), and for \(x\in C\), let \(a_s(x)=a_1(x)\) be the common source action. The decision-change set is \(D=\{x\in C:a_{12}(x)\ne a_s(x)\}\). We report both \(|D|/|X_{\mathrm{test}}|\) and \(|D|/|C|\), making the all-query and consensus-conditional rates explicit.

For task-defined utility \(u(x,a)\), define the signed utility loss relative to source consensus as \(\ell_u(x)=u(x,a_s(x))-u(x,a_{12}(x))\). A decision change is harmful when \(\ell_u(x)>0\), beneficial when \(\ell_u(x)<0\), and neutral when \(\ell_u(x)=0\). The changed-decision harmful fraction is \(h_D=|\{x\in D:\ell_u(x)>0\}|/|D|\). Dividing the same harmful-event count by \(|X_{\mathrm{test}}|\) or \(|C|\) gives the absolute harmful rates \(h_{\mathrm{all}}\) and \(h_C\), respectively. Together, these quantities characterize pairwise reversal, decision change, and task-defined utility loss.

\textbf{Stage linkage.} For a source-consensus query \(x\in C\), let \(a_s(x)\) be the common source action and \(a_{12}(x)\) the pooled action. The staged audit explicitly evaluates the pair \((a_s(x),a_{12}(x))\). A reported decision change is linked to a strict pairwise reversal only when both source gaps support \(a_s(x)\) beyond the source threshold and the pooled gap favors \(a_{12}(x)\) beyond \(\epsilon_{\mathrm{pair}}\). Under the reported decision-tie rule with \(\epsilon_{\mathrm{tie}}=10^{-6}\), this record-level check prevents near-ties from being inferred as reversals solely from the argmax. Whether the changed decision is harmful, beneficial, or neutral is then determined by the task-specific utility \(u\). Recorded-sex partitions are used solely as reliability stress tests and do not constitute deployment recommendations or justify attribute-based treatment.

\section{4 Gram Incompatibility and Preference Stability}

This section characterizes the preference consequences of inverse-Gram recomposition. In a pooled readout, source evidence is added while the source-specific inverse-Gram operators are replaced by the inverse Gram of the union. We establish that this recomposition can reverse a preference supported by both sources, identify scalar Gram alignment as an exact preservation regime, and derive a margin-dependent guarantee under approximate alignment.

\subsection{4.1 How Pooling Can Reverse a Shared Preference}

For an eligible query--action pair \(e\), let \(s=\operatorname{dir}_{\tau}(\Delta_1(e))=\operatorname{dir}_{\tau}(\Delta_2(e))\ne0\) denote the common source direction. A pairwise preference reversal occurs when the pooled margin points against this direction, \(s\Delta_{12}(e)<-\epsilon_{\mathrm{pair}}\).

The source and pooled margins apply different linear operators to return evidence. For a fixed action pair, write \(y_j=\Phi_j^\top d_j(a,b)\). Each source evaluates its evidence through \(G_j^{-1}y_j\), whereas pooling evaluates the combined evidence through \((G_1+G_2)^{-1}(y_1+y_2)\). For a fixed query representation and evidence vector, positive rescaling of a Gram matrix preserves its normalized direction and changes only the magnitude, not the sign, of the corresponding inverse-Gram margin. When the two source Gram matrices depart from a common direction beyond positive rescaling, the recomputed inverse geometry can rotate or reweight the combined evidence relative to the query representation. We refer to this departure as Gram incompatibility.

\textbf{Proposition 1 (Preference reversal under inverse-Gram recomposition).} There exist positive-definite source Gram matrices \(G_1,G_2\), source evidence vectors \(y_1,y_2\), and a query representation \(\phi(x)\) such that both source margins are positive while the pooled margin is negative.

A two-dimensional construction gives source margins \(0.465\) and \(0.635\), but a pooled margin of \(-0.134\). The construction establishes that source agreement alone does not determine the pooled ordering: recomputing the inverse Gram of the union can move the combined evidence across the preference boundary. The explicit matrices, their eigenvalues, and the three margin calculations are given in Appendix B.1.

\subsection{4.2 Exact Preservation under Scalar Gram Alignment}

An exact preservation regime arises when the two source Gram matrices share one positive-definite direction and differ only in scale.

\textbf{Proposition 2.} Suppose that there exist a positive-definite matrix \(G_0\) and positive scalars \(\kappa_1,\kappa_2\) such that \(G_j=\kappa_jG_0\) for both sources. Then

\begin{equation}
\Delta_{12}^{a,b}(x) = \frac{\kappa_{1}\Delta_{1}^{a,b}(x) + \kappa_{2}\Delta_{2}^{a,b}(x)}{\kappa_{1} + \kappa_{2}}.
\tag{7}\label{eq:b-08}
\end{equation}

Eq. (7) expresses the pooled margin as a convex combination of the two source margins, with positive weights \(\kappa_j/(\kappa_1+\kappa_2)\). A common weak ordering is therefore preserved, and the pooled ordering is strict whenever at least one source margin is strict. Positive scalar differences can change preference strength but preserve preference direction. The result follows by substituting \(G_j=\kappa_jG_0\) into the source and pooled margin formulas; the full derivation is given in Appendix B.2.

\subsection{4.3 Approximate Alignment and Margin Stability}

Exact scalar alignment provides the reference geometry for composition preservation. Under heterogeneous sources, the relevant deviation is the difference between Gram directions after source scale has been removed. Define the Frobenius-normalized Gram matrices and their normalized mismatch by

\begin{equation}
\begin{aligned}
\widetilde{G}_{j}
&= \frac{G_{j}}{\lVert G_{j}\rVert_{F}},\\
\delta_{\mathrm{norm}}(G_{1},G_{2})
&= \lVert \widetilde{G}_{1} - \widetilde{G}_{2}\rVert_{F}.
\end{aligned}
\tag{8}\label{eq:b-09}
\end{equation}

\textbf{Lemma 1.} For positive-semidefinite Gram matrices with positive Frobenius norm, \(\delta_{\mathrm{norm}}\) is symmetric, invariant to independent positive rescaling, and bounded above by \(\sqrt{2}\); the upper bound is strict for positive-definite matrices. Moreover, \(\delta_{\mathrm{norm}}(G_1,G_2)=0\) if and only if \(G_1\) and \(G_2\) are positive scalar multiples of one another.

The zero-mismatch regime therefore coincides exactly with the scalar-alignment condition in Proposition 2. The proof is given in Appendix B.3. Unlike scale-sensitive raw differences, \(\delta_{\mathrm{norm}}\) removes source magnitude and isolates the Gram-direction departure relevant to preference composition.

The mismatch-to-perturbation bridge converts this observable departure into the form required by the stability theorem. Choose
\[
\kappa_j=\|G_j\|_F,
\qquad
G_0=\frac{\widetilde G_1+\widetilde G_2}{2},
\qquad
E_j=G_j-\kappa_jG_0.
\]
Then
\[
\|E_j\|_2
\le
\|E_j\|_F
=
\frac{\kappa_j}{2}
\delta_{\mathrm{norm}}(G_1,G_2).
\]
Thus, normalized mismatch bounds departure from a common Gram direction up to positive scaling. Appendix B.4 gives the construction and the common perturbation level used below.

Approximate alignment replaces the exact convex combination in Eq. (7) with a weighted source margin minus a controlled perturbation. The following theorem quantifies this margin buffer.

\textbf{Theorem 1.} Let \(G_j=\kappa_jG_0+E_j\) for both sources, where \(G_0\) is positive definite, \(\kappa_j>0\), and \(\|E_j\|_2\le\epsilon\|G_0\|_2\). Let \(\kappa_{\min}=\min\{\kappa_1,\kappa_2\}\), \(\kappa_\Sigma=\kappa_1+\kappa_2\), \(\lambda_0=\lambda_{\min}(G_0)\), and \(y_j=\Phi_j^\top d_j(a,b)\). Assume

\begin{equation}
\varepsilon\lVert G_0\rVert_2
<
\lambda_0\min\left\{\frac{\kappa_{\min}}{2},\frac{\kappa_\Sigma}{4}\right\}.
\tag{9}\label{eq:b-10}
\end{equation}

Then

\begin{equation}
\begin{aligned}
\Delta_{12}^{a,b}(x)
&\geq
\frac{\kappa_{1}\Delta_{1}^{a,b}(x)
+ \kappa_{2}\Delta_{2}^{a,b}(x)}
{\kappa_\Sigma}\\
&\quad -
\frac{4\varepsilon\lVert\phi(x)\rVert_2\lVert G_0\rVert_2}
{\kappa_\Sigma\lambda_0^2}
\left(
\frac{\lVert y_1\rVert_2}{\kappa_1}
+\frac{\lVert y_2\rVert_2}{\kappa_2}
\right).
\end{aligned}
\tag{10}\label{eq:b-11}
\end{equation}

Eq. (9) keeps the source and pooled inverse-Gram operators in a stable perturbation regime. Eq. (10) then decomposes the pooled margin into the exact-alignment weighted source margin and an explicit mismatch-induced penalty. Preference preservation is guaranteed whenever the weighted source margin exceeds this penalty. Larger source margins and better conditioning enlarge the certified region, while larger Gram perturbations require a correspondingly larger margin buffer. The proof is given in Appendix B.5.

\textbf{Corollary 1 (Mismatch-to-stability certificate).} For \(G_1,G_2\succ0\), choose
\[
G_0=\frac{\widetilde G_1+\widetilde G_2}{2},
\qquad
\kappa_j=\|G_j\|_F,
\]
and define the conservative common perturbation level
\[
\epsilon_\delta
=
\frac{\delta_{\mathrm{norm}}(G_1,G_2)}{2\|G_0\|_2}
\max_{j\in\{1,2\}}\kappa_j.
\]
Then
\[
\|G_j-\kappa_jG_0\|_2
\le
\epsilon_\delta\|G_0\|_2
\]
for both sources. A directly checkable sufficient condition for Eq. (9) is
\[
\delta_{\mathrm{norm}}(G_1,G_2)
<
\lambda_0
\frac{\kappa_{\min}}{\max_j\kappa_j}.
\]
Whenever this condition holds and the weighted source margin in Eq. (10) exceeds the corresponding perturbation penalty, the pooled ordering is preserved.

Corollary 1 turns normalized mismatch into a conservative, query-specific preservation certificate. Mismatch supplies the pool-level geometric screen, while source margins, evidence-vector norms, source scales, and conditioning determine whether a particular query--action pair is certified. The condition is sufficient rather than necessary: an uncertified pair is not declared unstable but is passed to the empirical audit. This distinction links the screening workflow in Section 5 to the realized preference and decision outcomes measured in Section 6.

Together, these results establish a mechanism-to-stability chain: Gram incompatibility beyond positive scaling makes preference reversal possible, scalar alignment yields exact preservation, and approximate alignment yields a margin-dependent guarantee. The next section operationalizes the same geometry for candidate-pool screening and training-time adjustment.

\section{5 Geometry Screening and Training-Time Regularization}
This section operationalizes the Gram geometry characterized above. The normalized mismatch \(\delta_{\mathrm{norm}}\) measures departure between source Gram directions after scale removal and provides a pool-level audit-prioritization statistic. The certificate in Corollary 1 combines this screen with source margins, evidence norms, source scales, and conditioning; failure to satisfy it means only that a pair remains uncertified, not that it reverses.

Operationally, the workflow first computes \(G_1\), \(G_2\), and \(\delta_{\mathrm{norm}}\) to prioritize candidate pools. For an audited action pair, Corollary 1 is then used as a conservative sufficient check combining source margins, evidence norms, scales, and conditioning; uncertified pairs are passed to the strict empirical audit rather than declared unstable. Query-level aggregation finally separates decision changes from task-defined harmful, beneficial, and neutral outcomes.

During representation training, let
\[
\Psi_B=|B|^{-1}\Phi_B^\top\Phi_B
\]
denote a source-batch raw second-moment matrix, and let
\[
\Gamma_B=\Psi_B/\|\Psi_B\|_F
\]
denote its normalized Gram direction. We evaluate two geometry-oriented objectives in addition to standard weight decay. Gram whitening uses
\[
R_{\mathrm{white}}(B)=\|\Psi_B-I_H\|_F^2,
\]
moving each source stage toward a shared source-independent reference. Direct alignment uses
\[
R_{\mathrm{direct}}(B)=\|\Gamma_B-\Gamma_{\mathrm{low}}\|_F^2,
\]
where \(\Gamma_{\mathrm{low}}\) is the frozen terminal normalized Gram direction from the first, lower-demand source stage. Each objective is optimized as \(L_{\mathrm{task}}{+}\lambda_{\mathrm{reg}} R\). Direct alignment is applied only during the second training stage using the frozen first-stage Gram direction, so it is an order-dependent empirical baseline rather than a symmetric consequence of the preservation theorem. Weight decay serves as a non-geometric baseline. These objectives are evaluated as descriptive operating points because lower mismatch or higher strict A2 consistency need not imply lower downstream harm.

\section{6 Experiments}
The experiments move from mechanism to decision consequence and then to scope. The controlled construction first isolates Gram-incompatibility as a cause of strict A2 failure; the load-based bidding proxy then traces how strict reversals propagate to changed decisions and source-consensus-relative utility; the operating-point comparison checks whether geometry, accuracy, consistency, and harm move together; and the UCI Heart Disease and Default of Credit Card Clients audits \cite{janosi1989heart,yeh2009default} test how far mismatch-based screening transfers beyond the load proxy.

\begin{figure*}[t]
\centering
\begin{tabular}{@{}cc@{}}
\includegraphics[width=.485\textwidth]{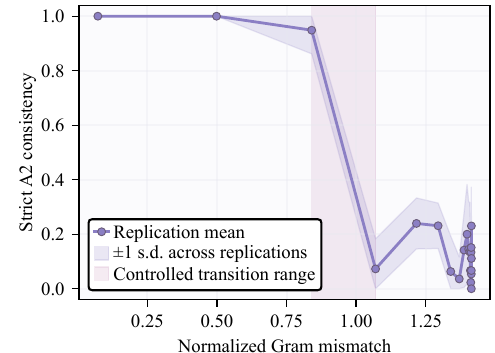} &
\includegraphics[width=.485\textwidth]{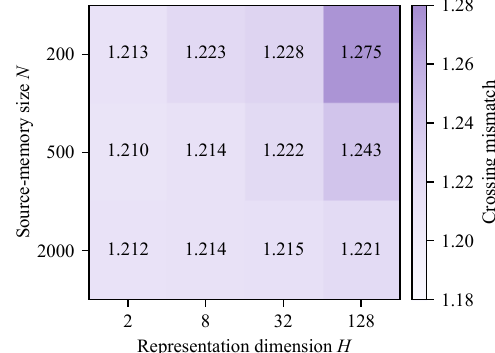} \\
\small (a) Main controlled sweep & \small (b) Robustness grid
\end{tabular}
\caption{Controlled strict A2 consistency as normalized Gram mismatch increases across the main sweep and the dimension--memory-size robustness grid. The repeated transition shows that inverse-Gram incompatibility can induce strict source-supported preference reversal without defining a universal cutoff. The consistency of the pattern across controlled settings provides strong mechanism-level support for the paper's geometric explanation.}
\label{fig:controlled_strict}
\end{figure*}

\begin{table*}[t]
\centering
\setlength{\tabcolsep}{5.0pt}
\begin{tabular}{p{0.23\textwidth}p{0.25\textwidth}cc}
\toprule
Audit quantity & Statistical unit & Count & Rate\\
\midrule
Strict pairwise reversal & Eligible query--action pair & \textbf{626 / 34,693} & \textbf{1.80\%}\\
Decision change & Source-consensus query & \textbf{110 / 9,837} & \textbf{1.12\%}\\
Harmful among changes & Changed decision & \textbf{59 / 110} & \textbf{53.64\%}\\
All-query harmful outcome & Test-query evaluation & \textbf{59 / 16,790} & \textbf{0.35\%}\\
Consensus-conditional harm & Source-consensus query & \textbf{59 / 9,837} & \textbf{0.60\%}\\
\bottomrule
\end{tabular}
\caption{Staged consequence audit for the canonical no-regularization load configuration, with counts pooled over five collision-free seed runs and the appropriate statistical unit and denominator for each quantity. The ledger separates formal pairwise failure from query-level decision change and shows that changed decisions can be either harmful or beneficial. This explicit denominator accounting provides a credible and conservative link from compositional failure to realized decision consequence.}
\label{tab:load_consequence_audit}
\end{table*}

\begin{figure*}[t]
\centering
\includegraphics[width=.98\textwidth]{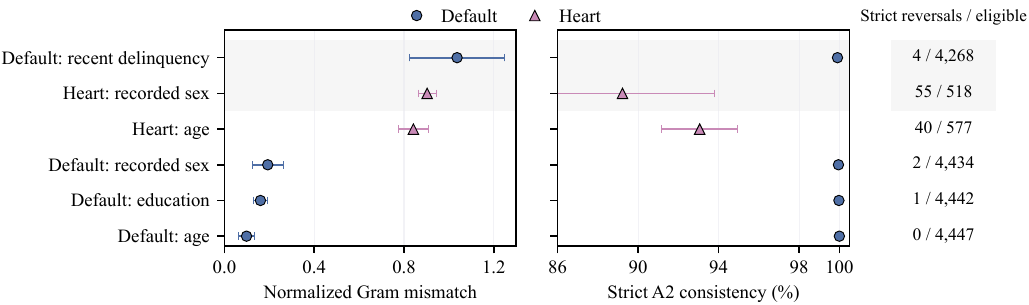}
\caption{Aligned cross-domain comparison of normalized Gram mismatch, strict A2 consistency, and strict-reversal incidence relative to the eligible-pair denominator for all six source partitions. The shared row order shows that comparable geometric mismatch can coexist with either near-perfect preservation or a measurable reversal regime. This contrast gives strong boundary evidence for using mismatch to prioritize audits rather than as a standalone failure predictor.}
\label{fig:cross_domain_aligned}
\end{figure*}

\begin{table*}[t]
\centering
\begin{tabular}{@{}lcccc@{}}
\toprule
Method & Acc. \(\uparrow\) & Strict A2 \(\uparrow\) & \(\delta_{\mathrm{norm}}\downarrow\) & \(h_{\mathrm{all}}(\%)\downarrow\)\\
\midrule
No regularization & 50.05\textpm{}10.13 & 98.16\textpm{}1.10 & 1.198\textpm{}0.056 & 0.35\textpm{}0.12\\
Gram whitening, 0.01 & \textbf{52.98\textpm{}3.73} & \textbf{98.83\textpm{}0.25} & \textbf{1.068\textpm{}0.051} & 0.40\textpm{}0.23\\
Direct alignment, 0.1 & 49.84\textpm{}10.04 & 98.20\textpm{}1.05 & 1.196\textpm{}0.056 & \textbf{0.31\textpm{}0.05}\\
Weight decay, \(10^{-2}\) & 50.98\textpm{}5.47 & 97.64\textpm{}1.19 & 1.168\textpm{}0.074 & 0.81\textpm{}0.90\\
\bottomrule
\end{tabular}
\caption{Representative five-seed accuracy--consistency--geometry--harm operating points under matched load returns. The best displayed point estimates occur at different configurations, supporting a multi-objective interpretation rather than a universally dominant objective.}
\label{tab:main_operating_points}
\end{table*}

\subsection{6.1 Experimental Protocol and Evidence Roles}
All canonical audits use source threshold \(\tau=0\) and numerical tolerance \(\epsilon_{\mathrm{pair}}=10^{-6}\), separately record strict reversals and pooled ties, compute the normalized Gram mismatch in Eq. (8), and use source constructions, proxy actions, statistical units, and evidence roles summarized in Appendix C. For the load proxy, each experimental seed initializes a separate return generator from master seed 20260725, persists that return realization, and reuses it across methods under the same seed. The medical and financial actions and returns are methodological audit proxies rather than clinical or credit recommendations, and values written as mean\textpm{}variation denote one standard deviation across seeds unless a 95\% interval is explicitly labeled.

\subsection{6.2 Controlled Mechanism Study}
This experiment isolates whether Gram-direction incompatibility can induce strict A2 failure under a controlled return protocol. Figure~\ref{fig:controlled_strict} shows that strict A2 consistency falls sharply as normalized mismatch grows. Across the evaluated main-sweep settings, the Pearson correlation between replication-mean normalized mismatch and replication-mean strict A2 consistency is -0.903, and the replication mean first falls below 0.5 at mismatch 1.070. Across the dimension--memory-size grid, the evaluated crossings range from 1.210 to 1.275, demonstrating a repeated transition without defining a universal cutoff.

The post-transition values are not monotone, and a few interpolation settings contain isolated pooled ties. The failures are nevertheless overwhelmingly strict directional reversals, so the mechanism result is not explained by numerical ties. Appendix D.6 reports the complete strict-reversal and pooled-tie ledger.

\subsection{6.3 Consequence Chain on the Load Proxy}
This experiment tests, under the canonical no-regularization load configuration, whether formal strict reversals propagate to changed pooled decisions and source-consensus-relative task utility. Table~\ref{tab:load_consequence_audit} separates the stages using the appropriate denominator for each quantity. Across five collision-free seed runs, 626 of 34,693 eligible query--action pairs reverse, yielding a pooled strict-reversal rate of 1.80\%, while 110 of 9,837 source-consensus queries change their pooled decision, a consensus-conditional change rate of 1.12\%.

The 110 changed decisions contain 59 outcomes that are harmful relative to the common source action under the persisted bidding-proxy utility and 51 beneficial outcomes, with no neutral cases. Pooling therefore does not uniformly worsen decisions, but strict compositional failures create a measurable nonzero harmful class: the changed-decision harmful fraction is \(h_D=53.64\%\), the all-query harmful rate is \(h_{\mathrm{all}}=0.35\%\), and the consensus-conditional harmful rate is \(h_C=0.60\%\). Each changed decision also passes the record-level strict-gap check for the pair consisting of the source-consensus and pooled actions, directly linking the decision-level change to a strict pairwise reversal. Appendix D.1 reports a representative source-consensus-oriented reversal. Source-margin filtering reduces retained failures but changes their consequence profile non-monotonically: harmful fractions remain near one half through threshold \(0.5\) and rise only in the sparse high-margin tail. Appendix D.2 reports the complete curve, denominators, and query-cluster bootstrap intervals.

\subsection{6.4 Geometry--Decision Operating Points}
This experiment tests whether geometry-oriented objectives improve the same downstream quantities or instead move the system among distinct operating points. Table~\ref{tab:main_operating_points} compares four representative configurations. Among the four representative rows in Table~\ref{tab:main_operating_points}, Gram whitening at \(\lambda_{\mathrm{reg}}=0.01\) has the highest displayed accuracy, strict A2 consistency, and lowest mismatch. Appendix D.3 reports the complete sweep, where the overall accuracy point estimate is slightly higher at whitening \(0.1\); this reinforces rather than weakens the conclusion that different metrics favor different operating points.

Absolute load-proxy accuracy varies substantially across the independently generated seed-specific return realizations. The matched design therefore compares methods within seed, where every configuration reuses the same returns, split, and test queries. Paired intervals are reported for \(\delta_{\mathrm{norm}}\), the intervention's geometric target, and \(h_{\mathrm{all}}\), the downstream consequence target; accuracy and strict A2 remain descriptive. Whitening at \(\lambda_{\mathrm{reg}}=0.01\) produces a paired mismatch reduction under both interval procedures, but its harmful-rate difference remains unresolved. The lower harmful-rate point estimate for direct alignment at \(\lambda_{\mathrm{reg}}=0.1\) is likewise unresolved, whereas the strongest whitening and direct-alignment penalties increase harm under both procedures. Lower mismatch is therefore measurable but is not a sufficient surrogate for lower decision harm. Appendix D.3 reports the complete operating map, and Appendix D.4 reports paired uncertainty.

\subsection{6.5 Cross-Domain Scope and Limitation}
This experiment tests whether normalized mismatch alone predicts realized strict preservation outside the controlled and load settings. Figure~\ref{fig:cross_domain_aligned} aligns the screening statistic and realized preservation for every evaluated medical and financial partition. Default recent delinquency and Heart recorded sex have comparable high mismatch, but their strict A2 consistency is 99.91\% and 89.22\%, with 4 and 55 strict reversals, respectively.

The contrast shows why mismatch is useful for audit prioritization but insufficient as a standalone predictor. Similar geometric incompatibility can coexist with either near-perfect preservation or a measurable reversal regime, consistent with the analytic certificate's additional dependence on source margins, evidence magnitude, source scale, and conditioning. Appendix D.5 reports the complete eligible-pair and reversal ledger.

\section{7 Conclusion}
This paper formalizes compositional reliability for pooled fixed-representation neural decisions. Inverse-Gram recomposition makes strict source-supported preference reversal possible; scalar alignment yields exact preservation, and approximate alignment gives a conservative margin-dependent certificate. The record-level load audit shows that every reported changed source-consensus decision is accompanied by a strict reversal of the corresponding source-versus-pooled action pair. Among these changed decisions, a measurable nonzero subset is harmful relative to the common source action under the task-defined proxy utility, while the remaining changes are beneficial; pooling is therefore not uniformly harmful. Cross-domain audits show that comparable mismatch can correspond to either near-perfect preservation or a measurable reversal regime, clarifying both the value and limitation of mismatch-based screening. The operating-point results further show that geometry, strict consistency, accuracy, and harm need not rank training objectives identically.

The present analysis is limited to guarantees conditional on a frozen representation at readout-fitting time, linear OLS/ridge heads, proxy actions and returns, five-seed paired comparisons, one fixed source order for direct alignment, and sparse high-margin tails. Future work should extend the framework to multiple memories, nonlinear readouts, scalable Gram approximations, outcome-grounded returns, and direct empirical evaluation of analytic-certificate coverage.

\bibliographystyle{unsrtnat}
\bibliography{refs}

\clearpage
\twocolumn[
\supplementnotation
]

\renewcommand{\thesection}{\Alph{section}}
\section{Appendix A Notation}
Table A1 provides the consolidated notation used throughout the theoretical analysis, decision audit, representation objectives, and experiments.

\setcounter{figure}{0}\setcounter{table}{1}\renewcommand{\thefigure}{D\arabic{figure}}\renewcommand{\thetable}{D\arabic{table}}
\begin{figure*}[t]\centering\begin{tabular}{@{}cc@{}}
\includegraphics[width=.485\textwidth]{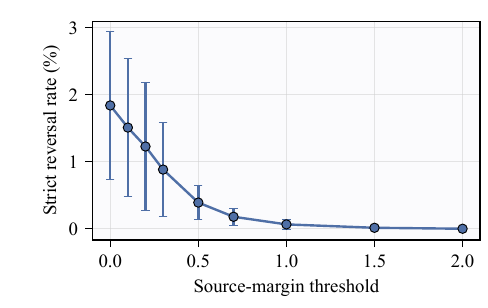} & \includegraphics[width=.485\textwidth]{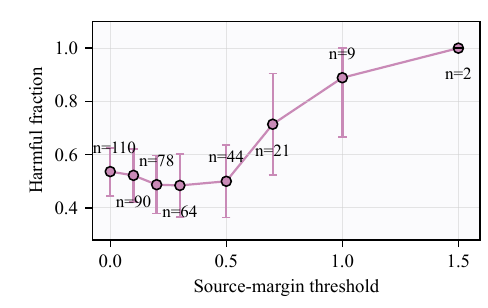}\\
\small (a) Strict reversal rate & \small (b) Harmful fraction
\end{tabular}
\caption{Five-seed strict-reversal rates and changed-decision harmful fractions across source-margin thresholds, with seed-level variation in panel (a) and chronological-query cluster-bootstrap intervals in panel (b). Filtering progressively reduces the retained failure set, while the apparent consequence shift is concentrated in the sparse high-margin tail. The diagnostic supports disciplined sensitivity analysis and guards against treating a margin threshold as a universal operating law.}\label{fig:threshold_diagnostic}\end{figure*}
\begin{table*}[t]\centering\small
\begin{tabular}{rrrrrrl}\toprule
Threshold & Changed & Distinct query clusters & Harmful & Beneficial & \(h_D\) & Cluster-bootstrap 95\% CI\\\midrule
0.0 & 110 & 98 & 59 & 51 & 0.536 & [0.444, 0.625]\\
0.1 & 90 & 85 & 47 & 43 & 0.522 & [0.422, 0.622]\\
0.2 & 78 & 75 & 38 & 40 & 0.487 & [0.380, 0.597]\\
0.3 & 64 & 62 & 31 & 33 & 0.484 & [0.365, 0.603]\\
0.5 & 44 & 44 & 22 & 22 & 0.500 & [0.364, 0.636]\\
0.7 & 21 & 21 & 15 & 6 & 0.714 & [0.524, 0.905]\\
1.0 & 9 & 9 & 8 & 1 & 0.889 & [0.667, 1.000]\\
1.5 & 2 & 2 & 2 & 0 & 1.000 & [1.000, 1.000]\\
2.0 & 0 & 0 & 0 & 0 & -- & --\\\bottomrule
\end{tabular}
\caption{Threshold-specific changed-decision counts, distinct chronological-query clusters, outcome composition, and cluster-bootstrap intervals. The apparent rise in harmful fraction emerges only after the retained denominator becomes sparse. Complete denominator and uncertainty reporting makes this limitation transparent and supports a cautious descriptive interpretation.}\label{tab:threshold_diagnostic}\end{table*}
\begin{figure*}[t]\centering\includegraphics[width=.95\textwidth]{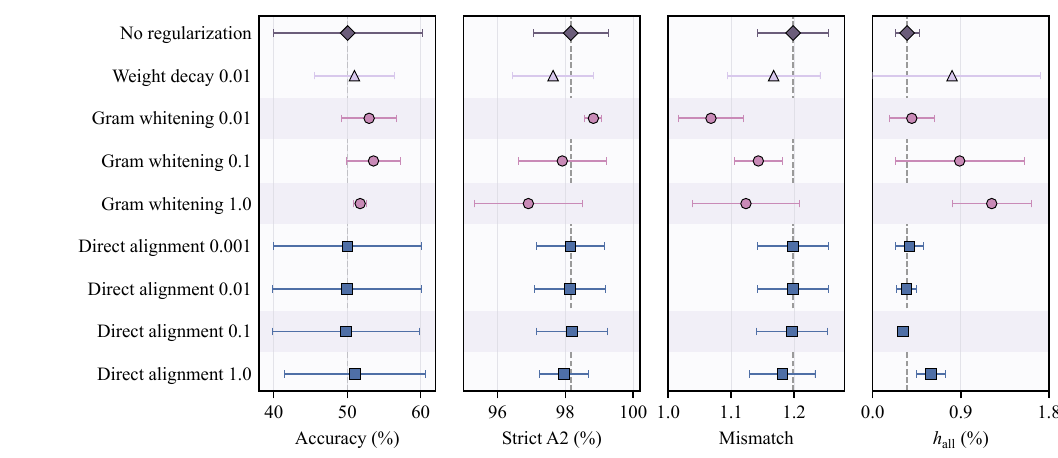}
\caption{Aligned operating-point profiles showing five-seed means and one-standard-deviation intervals for accuracy, strict A2 consistency, normalized Gram mismatch, and all-query harmful rate across all nine configurations. No configuration dominates all four panels, and geometry-oriented objectives move different evaluation criteria in different directions. The aligned profiles provide clear visual support for a multi-objective interpretation rather than a single universally best objective.}\label{fig:operating_map}\end{figure*}
\begin{table*}[t]
\centering\footnotesize\setlength{\tabcolsep}{2.2pt}
\begin{tabular}{llrrrrrr}
\toprule
Objective & \(\lambda_{\mathrm{reg}}\) & Acc. (\%) & Strict A2 (\%) & \(\delta_{\mathrm{norm}}\) & All-query change (\%) & \(h_D\) (\%) & \(h_{\mathrm{all}}\) (\%)\\
\midrule
None & 0 & 50.05\textpm{}10.13 & 98.16\textpm{}1.10 & 1.1984\textpm{}0.0562 & 0.66\textpm{}0.25 & 55.19\textpm{}6.96 & 0.35\textpm{}0.12\\
Weight decay & 0.01 & 50.98\textpm{}5.47 & 97.64\textpm{}1.19 & 1.1675\textpm{}0.0736 & 1.48\textpm{}1.48 & 55.38\textpm{}16.08 & 0.81\textpm{}0.90\\
Gram whitening & 0.01 & 52.98\textpm{}3.73 & 98.83\textpm{}0.25 & 1.0680\textpm{}0.0514 & 0.74\textpm{}0.31 & 50.47\textpm{}13.30 & 0.40\textpm{}0.23\\
Gram whitening & 0.1 & 53.57\textpm{}3.70 & 97.91\textpm{}1.30 & 1.1432\textpm{}0.0385 & 1.63\textpm{}1.33 & 56.91\textpm{}10.65 & 0.89\textpm{}0.66\\
Gram whitening & 1.0 & 51.76\textpm{}0.90 & 96.91\textpm{}1.60 & 1.1235\textpm{}0.0844 & 1.98\textpm{}0.39 & 60.12\textpm{}10.91 & 1.22\textpm{}0.40\\
Direct alignment & 0.001 & 50.03\textpm{}10.13 & 98.15\textpm{}1.01 & 1.1985\textpm{}0.0563 & 0.71\textpm{}0.28 & 53.32\textpm{}7.53 & 0.38\textpm{}0.14\\
Direct alignment & 0.01 & 49.93\textpm{}10.11 & 98.13\textpm{}1.05 & 1.1982\textpm{}0.0563 & 0.66\textpm{}0.22 & 54.55\textpm{}10.48 & 0.35\textpm{}0.10\\
Direct alignment & 0.1 & 49.84\textpm{}10.04 & 98.20\textpm{}1.05 & 1.1964\textpm{}0.0561 & 0.67\textpm{}0.15 & 47.62\textpm{}9.20 & 0.31\textpm{}0.05\\
Direct alignment & 1.0 & 51.07\textpm{}9.62 & 97.96\textpm{}0.71 & 1.1811\textpm{}0.0521 & 1.08\textpm{}0.33 & 55.93\textpm{}4.77 & 0.60\textpm{}0.15\\
\bottomrule
\end{tabular}
\caption{Complete five-seed operating-point sweep under matched collision-free load returns. The best displayed point estimates for accuracy, strict consistency, normalized mismatch, conditional harm, and all-query harm occur at different configurations. The complete ledger supports transparent multi-objective comparison without implying statistically resolved or universal dominance.}
\label{tab:operating_full}
\end{table*}

\begin{table*}[t]
\centering\footnotesize\setlength{\tabcolsep}{2.0pt}
\begin{tabular}{llrrrl}
\toprule
Method & \(\lambda\) & Mean diff. (pp) & Paired \(t\) 95\% CI & Bootstrap 95\% CI & Reading\\
\midrule
Direct alignment & 0.001 & +0.024 & [-0.047, +0.095] & [-0.012, +0.071] & unresolved\\
Direct alignment & 0.01 & -0.006 & [-0.061, +0.049] & [-0.036, +0.030] & unresolved\\
Direct alignment & 0.1 & -0.042 & [-0.166, +0.083] & [-0.125, +0.036] & unresolved\\
Direct alignment & 1.0 & \textbf{+0.244} & \textbf{[+0.118, +0.371]} & \textbf{[+0.173, +0.316]} & higher harm\\
Gram whitening & 0.01 & +0.048 & [-0.144, +0.239] & [-0.066, +0.179] & unresolved\\
Gram whitening & 0.1 & +0.536 & [-0.156, +1.228] & [+0.107, +0.965] & interval-sensitive\\
Gram whitening & 1.0 & \textbf{+0.864} & \textbf{[+0.405, +1.322]} & \textbf{[+0.530, +1.126]} & higher harm\\
Weight decay & 0.01 & +0.459 & [-0.799, +1.716] & [-0.048, +1.370] & unresolved\\
\midrule
\multicolumn{6}{l}{Panel B: selected \(\Delta\delta_{\mathrm{norm}}\)}\\
Gram whitening & 0.01 & \textbf{-0.1304} & \textbf{[-0.2560, -0.0049]} & \textbf{[-0.1884, -0.0396]} & lower mismatch\\
Direct alignment & 1.0 & \textbf{-0.0174} & \textbf{[-0.0316, -0.0031]} & \textbf{[-0.0269, -0.0097]} & lower mismatch\\
Weight decay & 0.01 & \textbf{-0.0309} & \textbf{[-0.0613, -0.00055]} & \textbf{[-0.0495, -0.0127]} & lower mismatch\\
Direct alignment & 0.1 & -0.0020 & [-0.0047, +0.00069] & [-0.0039, -0.00081] & interval-sensitive\\
\bottomrule
\end{tabular}
\caption{Paired five-seed differences relative to no regularization for all-query harmful rate and selected normalized-mismatch comparisons. Several objectives improve Gram compatibility without establishing a corresponding reduction in task-defined harm. This paired evidence strengthens the conclusion that geometric improvement is not a reliable surrogate for downstream decision utility.}
\label{tab:paired_operating}
\end{table*}
\begin{table*}[t]
\centering\small
\begin{tabular}{llrrrr}
\toprule
Domain & Partition & Eligible pairs & Strict reversals & \(\delta_{\mathrm{norm}}\) & Strict A2 consistency\\
\midrule
Default Credit & Age & 4,447 & 0 & 0.099\textpm{}0.035 & 1.0000\textpm{}0.0000\\
Default Credit & Education & 4,442 & 1 & 0.160\textpm{}0.030 & 0.9998\textpm{}0.0005\\
Default Credit & Recent delinquency & 4,268 & 4 & 1.036\textpm{}0.211 & 0.9991\textpm{}0.0015\\
Default Credit & Recorded sex & 4,434 & 2 & 0.193\textpm{}0.068 & 0.9995\textpm{}0.0006\\
Heart Disease & Age & 577 & 40 & 0.842\textpm{}0.066 & 0.9306\textpm{}0.0188\\
Heart Disease & Recorded sex & 518 & 55 & 0.903\textpm{}0.040 & 0.8922\textpm{}0.0458\\
\bottomrule
\end{tabular}
\caption{Complete five-seed cross-domain strict composition audit with pooled eligible-pair and strict-reversal counts and seed-level mismatch and consistency summaries. Comparable high-mismatch partitions occupy sharply different realized preservation regimes. This full ledger provides strong cross-domain support for the scope and limitation of mismatch-based screening.}
\label{tab:cross_domain_full}
\end{table*}

\begin{table*}[t]
\centering\small
\begin{tabular}{rrrrrrrr}
\toprule
\(H\) & \(N\) & \(\alpha\) & Eligible pairs (sum) & Mean \(\delta_{\mathrm{norm}}\) & Mean strict A2 & Strict reversals (sum) & Pooled ties (sum)\\
\midrule
2 & 600 & 0.00 & 10,712 & 0.072 & 1.000 & 0 & 0\\
2 & 600 & 0.05 & 10,996 & 0.498 & 1.000 & 0 & 0\\
2 & 600 & 0.10 & 9,635 & 0.840 & 0.949 & 517 & 1\\
2 & 600 & 0.15 & 9,369 & 1.070 & 0.073 & 8,632 & 0\\
2 & 600 & 0.20 & 10,638 & 1.217 & 0.240 & 7,991 & 0\\
2 & 600 & 0.25 & 10,601 & 1.295 & 0.231 & 8,072 & 1\\
2 & 600 & 0.30 & 8,696 & 1.340 & 0.063 & 7,992 & 0\\
2 & 600 & 0.50 & 10,088 & 1.405 & 0.145 & 8,386 & 0\\
2 & 600 & 0.90 & 8,785 & 1.414 & 0.000 & 8,785 & 0\\
2 & 600 & 1.00 & 10,391 & 1.414 & 0.111 & 9,051 & 7\\
\bottomrule
\end{tabular}
\caption{Controlled main-sweep ledger with pooled eligible-pair, strict-reversal, and tie counts together with replication-mean mismatch and strict A2 consistency. Strict directional reversals drive the preservation collapse, whereas pooled ties remain isolated and negligible. The detailed accounting supports the inverse-Gram mechanism while making the two aggregation levels explicit.}
\label{tab:controlled_full}
\end{table*}
\setcounter{table}{0}\renewcommand{\thetable}{C\arabic{table}}\renewcommand{\thefigure}{D\arabic{figure}}

\section{Appendix B Proofs and Derivations}
\subsection{B.1 Verified Reversal Construction for Proposition 1}
Let
\begin{equation}
\begin{gathered}
G_1=\begin{bmatrix}4.022&1.044\\1.044&0.631\end{bmatrix},\quad
G_2=\begin{bmatrix}0.644&-0.578\\-0.578&2.451\end{bmatrix},\\
\phi(x)=\begin{bmatrix}1.442\\1.021\end{bmatrix},\quad
 y_1=\begin{bmatrix}-0.750\\0.214\end{bmatrix},\quad
 y_2=\begin{bmatrix}0.267\\-0.209\end{bmatrix}.
\end{gathered}
\tag{B1}
\end{equation}
The eigenvalues of \(G_1\) are 0.335 and 4.318, and those of \(G_2\) are 0.475 and 2.620. Direct substitution gives \(\phi(x)^\top G_1^{-1}y_1=0.465\), \(\phi(x)^\top G_2^{-1}y_2=0.635\), and \(\phi(x)^\top(G_1+G_2)^{-1}(y_1+y_2)=-0.134\). Both sources support the same strict ordering, whereas the pooled readout reverses it. The construction is realizable by finite source memories: for each \(G_j\succ0\), choose full-column-rank \(\Phi_j\) with \(\Phi_j^\top\Phi_j=G_j\); then \(d_j=\Phi_jG_j^{-1}y_j\) gives \(\Phi_j^\top d_j=y_j\).

\subsection{B.2 Exact Preservation under Scalar Gram Alignment}
If \(G_j=\kappa_jG_0\), then \((G_1+G_2)^{-1}=(\kappa_1+\kappa_2)^{-1}G_0^{-1}\) and \(G_j^{-1}=\kappa_j^{-1}G_0^{-1}\). With \(u_j=\phi(x)^\top G_0^{-1}\Phi_j^\top d_j(a,b)\),
\[
\Delta_{12}^{a,b}(x)=\frac{u_1+u_2}{\kappa_1+\kappa_2}=\frac{\kappa_1\Delta_1^{a,b}(x)+\kappa_2\Delta_2^{a,b}(x)}{\kappa_1+\kappa_2}.
\tag{B2}
\]
The pooled margin is a convex combination of the source margins, proving weak and strict A2 preservation.

\subsection{B.3 Proof of Lemma 1}
For positive-semidefinite matrices, \(\|\widetilde G_j\|_F=1\) and \(\langle\widetilde G_1,\widetilde G_2\rangle_F\ge0\). Hence \(\delta_{\mathrm{norm}}^2=2-2\langle\widetilde G_1,\widetilde G_2\rangle_F\le2\). Zero mismatch holds exactly when the normalized matrices are equal, equivalently when the raw matrices are positive scalar multiples. If both Gram matrices are positive definite, then \(\langle \widetilde G_1,\widetilde G_2\rangle_F=\operatorname{tr}(\widetilde G_1\widetilde G_2)>0\), so the upper bound is strict: \(\delta_{\mathrm{norm}}(G_1,G_2)<\sqrt{2}\).

\subsection{B.4 Mismatch-to-Perturbation Bridge and Corollary 1}
Let \(\widetilde G_j=G_j/\|G_j\|_F\), \(\kappa_j=\|G_j\|_F\), \(G_0=(\widetilde G_1+\widetilde G_2)/2\), and \(E_j=G_j-\kappa_jG_0\). Because \(G_1,G_2\succ0\), their normalized average satisfies \(G_0\succ0\). Moreover,
\[
E_1=\frac{\kappa_1}{2}(\widetilde G_1-\widetilde G_2),\qquad
E_2=\frac{\kappa_2}{2}(\widetilde G_2-\widetilde G_1),
\]
so
\[
\|E_j\|_2\le\|E_j\|_F=\frac{\kappa_j}{2}\delta_{\mathrm{norm}}(G_1,G_2).
\tag{B3}
\]
Define \(\epsilon_\delta=\delta_{\mathrm{norm}}(G_1,G_2)\max_j\kappa_j/(2\|G_0\|_2)\). Then \(\|E_j\|_2\le\epsilon_\delta\|G_0\|_2\) for both sources. Since \(\kappa_\Sigma\ge2\kappa_{\min}\), \(\min\{\kappa_{\min}/2,\kappa_\Sigma/4\}=\kappa_{\min}/2\). Thus Eq.~(9) is implied by \(\delta_{\mathrm{norm}}(G_1,G_2)<\lambda_0\kappa_{\min}/\kappa_{\max}\). Substituting \(\epsilon_\delta\) into Eq.~(10) gives Corollary 1.

\subsection{B.5 Approximate-Alignment Bound for Theorem 1}
Let \(G_\Sigma=G_1+G_2\), \(H_j=(\kappa_\Sigma/\kappa_j)G_j\), and \(\kappa_\Sigma=\kappa_1+\kappa_2\). Then \(H_j^{-1}=\kappa_jG_j^{-1}/\kappa_\Sigma\), so \(\phi(x)^\top H_j^{-1}y_j=\kappa_j\Delta_j^{a,b}(x)/\kappa_\Sigma\). The resolvent identity gives \(G_\Sigma^{-1}-H_j^{-1}=G_\Sigma^{-1}(H_j-G_\Sigma)H_j^{-1}\). Under Eq.~(9), \(\|G_\Sigma^{-1}\|_2\le2/(\kappa_\Sigma\lambda_0)\), \(\|H_j^{-1}\|_2\le2/(\kappa_\Sigma\lambda_0)\), and \(\|H_j-G_\Sigma\|_2\le\epsilon\|G_0\|_2\kappa_\Sigma/\kappa_j\). Hence
\[
\|G_\Sigma^{-1}-H_j^{-1}\|_2\le\frac{4\epsilon\|G_0\|_2}{\kappa_\Sigma\kappa_j\lambda_0^2}.
\tag{B4}
\]
Cauchy--Schwarz yields \(|\phi(x)^\top(G_\Sigma^{-1}-H_j^{-1})y_j|\le4\epsilon\|\phi(x)\|_2\|G_0\|_2\|y_j\|_2/(\kappa_\Sigma\kappa_j\lambda_0^2)\). For each source,
\[
\phi(x)^\top G_\Sigma^{-1}y_j
\ge
\frac{\kappa_j\Delta_j^{a,b}(x)}{\kappa_\Sigma}
-
\frac{4\epsilon\|\phi(x)\|_2\|G_0\|_2\|y_j\|_2}
{\kappa_\Sigma\kappa_j\lambda_0^2}.
\]
Summing this inequality over \(j=1,2\) and using \(\Delta_{12}^{a,b}(x)=\phi(x)^\top G_\Sigma^{-1}(y_1+y_2)\) gives Eq.~(10).

\subsection{B.6 Ridge-Stabilized Margin Perturbation}
For \(G\succ0\), let \(m_0=\phi(x)^\top G^{-1}y\), \(m_{\lambda_{\mathrm{ridge}}}=\phi(x)^\top(G+\lambda_{\mathrm{ridge}}I)^{-1}y\), and \(\mu=\lambda_{\min}(G)>0\). The resolvent identity gives
\[
(G+\lambda_{\mathrm{ridge}}I)^{-1}-G^{-1}=-(G+\lambda_{\mathrm{ridge}}I)^{-1}\lambda_{\mathrm{ridge}}IG^{-1}.
\]
Therefore
\[
|m_{\lambda_{\mathrm{ridge}}}-m_0|
\le
\frac{\lambda_{\mathrm{ridge}}\|\phi(x)\|_2\|y\|_2}{\mu(\mu+\lambda_{\mathrm{ridge}})}.
\tag{B5}
\]
If \(|m_0|\) exceeds this bound, then the ridge-stabilized and OLS margins have the same sign.

\section{Appendix C Experimental Protocol and Reproducibility}
\setcounter{table}{0}\renewcommand{\thetable}{C\arabic{table}}

\begin{table*}[t]
\centering\footnotesize\setlength{\tabcolsep}{3pt}
\begin{tabularx}{\textwidth}{>{\raggedright\arraybackslash}p{0.12\textwidth}>{\raggedright\arraybackslash}X>{\raggedright\arraybackslash}p{0.20\textwidth}>{\raggedright\arraybackslash}p{0.17\textwidth}>{\raggedright\arraybackslash}p{0.20\textwidth}}
\toprule Experiment & Source construction & Actions/returns & Primary unit & Evidence role\\\midrule
Controlled & Synthetic Gram-direction sweep & Three-action fixed protocol & Eligible pair & Mechanism and transition robustness\\
Load proxy & Training-defined low/high load memories & Three bidding-proxy returns & Pair, consensus query, outcome & Full consequence chain\\
Heart Disease & Age or recorded-sex partitions & Triage-intensity proxy & Eligible pair & Cross-domain scope\\
Default Credit & Age, education, recorded sex, recent delinquency & Exposure proxy & Eligible pair & Cross-domain boundary evidence\\
Operating points & Matched load runs across objectives & Same per-seed returns & Matched seed; pair/query summaries within seed & Accuracy--consistency--geometry--harm tradeoff\\\bottomrule
\end{tabularx}
\caption{Decision-proxy domains, source constructions, statistical units, and evidential roles used in the reported experiments. The table makes the division of labor among mechanism, consequence-chain, operating-point, and cross-domain evaluations explicit. This organization strengthens the experimental narrative by showing that each study supplies a distinct and complementary form of evidence.}
\label{tab:protocol_roles}
\end{table*}

\subsection{C.1 Unified Strict A2 Audit}
All canonical audits use source threshold \(\tau=0\) and numerical tolerance \(\epsilon_{\mathrm{pair}}=10^{-6}\). For each test query, the audit enumerates all \(\binom{K}{2}\) unordered action pairs. Eligibility requires both source margins to have the same nonzero strict direction; Appendix D.2 varies \(\tau\) as a sensitivity diagnostic. Strict reversals and pooled-tie strictness failures are exported separately. The audit therefore evaluates the strict--strict subset of the A2 antecedent rather than weak--strict cases involving a source tie.

\subsection{C.2 Controlled Construction and Robustness Grid}
The main controlled sweep uses \(\alpha\in\{0,0.05,0.10,\ldots,0.95,1.00\}\). At interpolation value \(\alpha\), the two source feature scales are \((e^{-2.5\alpha},e^{2.5\alpha})\) and \((e^{2.5\alpha},e^{-2.5\alpha})\). The return-difference directions are \(d_1=(3,-1)\) and \(d_2=(-1,3)\), with \(\phi(x)=x\). Independent Gaussian return noise with standard deviation 0.05 is used in the main sweep. The main sweep uses two source memories of 600 cases, three actions, 800 accepted queries per replication, five replications per setting, and readout ridge \(10^{-6}\). The robustness grid uses \(H\in\{2,8,32,128\}\), \(N\in\{200,500,2000\}\), 300 accepted queries, interpolation values \(\alpha\in\{0,0.1,\ldots,1.0\}\), return-noise standard deviation 0.3, and five replications per cell. The reported curve uses unweighted replication means and one standard deviation; the transition is the first evaluated mismatch at which the replication-mean strict A2 consistency is below 0.5.

\subsection{C.3 Decision-Proxy Datasets and Source Partitions}
Heart Disease uses the 297 complete Cleveland records and seeds 42--46 with random unstratified 80/20 splits, yielding 237 training and 60 test cases per seed. Continuous variables are standardized using training-only statistics. Source memories are defined within the training split by median age or recorded sex. Diagnosis class \(c\in\{0,\ldots,4\}\) maps to the proxy returns \((1.0,0.3,-0.2)\), \((0.2,0.9,0.5)\), \((-0.1,1.0,0.7)\), \((-0.3,0.6,1.0)\), and \((-0.5,0.4,1.1)\). Default Credit uses seeds 42--46 with random unstratified 80/20 splits. The identifier and target columns are removed, published categorical codes are retained as numeric inputs, and all standardization and partition thresholds use training data only. The source partitions use median age, education, recorded sex, or recent delinquency. For default indicator \(c\in\{0,1\}\), the three exposure-proxy returns are \((0.5(1-c),1-2c,2-5c)\). These proxy actions are methodological audit constructs rather than clinical or credit recommendations; recorded-sex partitions are used solely as reliability stress tests and do not constitute deployment recommendations or justify attribute-based treatment.

\subsection{C.4 Collision-Free Load Returns and Proxy Utility}
For each experimental seed, the ordered load dataset contains 16,790 row--slot queries and is split chronologically into 13,432 training and 3,358 test queries. The load consequence totals pool the 3,358 test-query evaluations from each of five seeds, giving 16,790 test-query evaluations in Table 1; this pooled test total is numerically equal to, but conceptually distinct from, the single-seed full-dataset size.

The low- and high-demand source memories are defined using a training-only load threshold. For ordered row--slot index \(t\), with \(h\in\{2,\ldots,47\}\), the feature vector is \(x_t=(L_{i,h},L_{i,h-2},\sin(2\pi h/48),\cos(2\pi h/48))\). The three bid levels are 25, 45, and 70 USD/MWh. The clipped normalized load is \(\bar L_{i,h}=\operatorname{clip}((L_{i,h}-700)/600,-0.2,1.5)\), and the deterministic base-price component is \(z_{i,h}=28+18\bar L_{i,h}+22\bar L_{i,h}^2\). The reported generator applies peak-slot and ramp adjustments, Gaussian price noise, and a rare spike process in fixed row--slot traversal order.

For action \(a\) with bid \(b_a\), task utility is the persisted bidding-proxy return \(u(x,a)=r(a,m)\), where \(r(a,m)=(m-b_a)(1+0.6a)-18\) if \(m\ge b_a\) and \(r(a,m)=-3(a+1)\) otherwise. A changed decision is harmful only relative to the common source action under this utility, as defined by \(\ell_u(x)\); it is not an absolute statement that either action is globally optimal. Utility consequences are classified from the persisted proxy returns without an additional numerical tolerance; exact zero is recorded as neutral. For each experimental seed, a separate pseudorandom generator is initialized from \(\operatorname{SeedSequence}([20260725,\text{experimental seed}])\), and the complete price and return arrays are persisted as seed-specific NPZ files.

\subsection{C.5 Training Objectives and Statistical Units}
Within each method and seed, representation training produces one trained frozen map \(\phi\). The source-specific and pooled readouts, Gram matrices, mismatch values, and strict audits are then computed using that same map; the representation is not refitted separately for the source and pooled OLS heads. The temporary action-value head is trained using \(L_{\mathrm{task}}=|B|^{-1}\sum_{(x,r)\in B}\|f_\theta(x)-r(x)\|_2^2\), then discarded before action-specific OLS/ridge readouts are fitted.

The load representation uses one 32-unit ReLU hidden layer, Heart Disease uses a 32--16 architecture, and Default Credit uses a 64--32--16 architecture. Representation training uses Adam with \((\beta_1,\beta_2)=(0.9,0.999)\), learning rate \(10^{-3}\), batch size 128, and 80 epochs per stage. The objectives use \(L_{\mathrm{task}}+\lambda_{\mathrm{reg}}R\), with \(R_{\mathrm{white}}(B)=\|\Psi_B-I_H\|_F^2\) and \(R_{\mathrm{direct}}(B)=\|\Gamma_B-\Gamma_{\mathrm{low}}\|_F^2\). Direct alignment is applied during the second, higher-demand stage using the frozen terminal normalized Gram direction from the first, lower-demand stage, so it is an order-dependent empirical baseline.

For the load proxy, decision accuracy is \(\mathrm{Acc}=|\mathcal X_{\mathrm{test}}|^{-1}\sum_{x\in\mathcal X_{\mathrm{test}}}\mathbf 1\{a_{12}(x)=\arg\max_a u(x,a)\}\), with the same deterministic tie rule used by the decision audit. Operating-point entries in Table D3 are unweighted means and one standard deviation over five matched seeds. The All-query change column is \(|D|/|\mathcal X_{\mathrm{test}}|\), whereas the consensus-conditional change rate reported in Table 1 is \(|D|/|C|\). The harmful-fraction column is \(h_D\), and the last column is \(h_{\mathrm{all}}\). Paired operating-point comparisons treat the five matched seeds as statistical units and are interpreted as descriptive paired uncertainty rather than multiplicity-adjusted superiority claims.

\subsection{C.6 Reproducibility Artifacts}
The released artifact retains the seed-specific load returns, fixed train--test splits, per-seed audit outputs, summary tables, plotting inputs, and the runner needed to reproduce the reported tables and figures. Its manifest records the correspondence between experimental outputs and manuscript artifacts. Table~\ref{tab:protocol_roles} summarizes the source construction, proxy action space, statistical unit, and evidential role of each reported experiment.
\section{Appendix D Additional Experimental Results}
\renewcommand{\thetable}{D\arabic{table}}\renewcommand{\thefigure}{D\arabic{figure}}\setcounter{table}{0}
\subsection{D.1 Representative Query-Level Reversal}
Table~\ref{tab:representative_reversal} reports a representative source-consensus reversal. The common source action is 0 and the pooled action is 1; the source-consensus-oriented margins are positive in both sources but negative after pooling.
\begin{table}[t]
\centering
\scriptsize
\setlength{\tabcolsep}{3pt}
\begin{tabular}{@{}rrrrrrr@{}}
\toprule
Seed & Query & Source & Pooled & \(\Delta_1\) & \(\Delta_2\) & \(\Delta_{12}\)\\
\midrule
46 & 2138 & 0 & 1 & +1.800 & +1.730 & -0.086\\
\bottomrule
\end{tabular}
\caption{Representative reported source-consensus reversal. Both source readouts favor the common source action, whereas the pooled readout favors the alternative action. The sign pattern links the decision change to a strict pairwise reversal rather than a numerical near-tie.}
\label{tab:representative_reversal}
\end{table}
\subsection{D.2 Source-Margin Threshold Diagnostic} Figure~\ref{fig:threshold_diagnostic} and Table~\ref{tab:threshold_diagnostic} describe the retained strict-reversal frequency and changed-decision consequence profile across source-margin thresholds. The degenerate interval at threshold 1.5 reflects two harmful records in two clusters and should not be interpreted as high statistical precision.
\subsection{D.3 Full Operating-Point Sweep} Figure~\ref{fig:operating_map} compares all nine reported configurations across accuracy, strict A2 consistency, normalized Gram mismatch, and all-query harmful rate using aligned metric profiles. The All-query change column in Table~\ref{tab:operating_full} uses \(|D|/|\mathcal X_{\mathrm{test}}|\), whereas Main Table 1 reports the consensus-conditional rate \(|D|/|C|\); the two values use different denominators by design. The shared row order shows that no configuration dominates all four quantities: Gram whitening at \(\lambda_{\mathrm{reg}}=0.01\) has the strongest displayed geometry--accuracy--consistency profile, while direct alignment at \(\lambda_{\mathrm{reg}}=0.1\) has the lowest harmful-rate point estimate. Exact five-seed means and standard deviations are reported in Table~\ref{tab:operating_full}, while Table~\ref{tab:paired_operating} reports paired seed-level differences relative to no regularization.

\subsection{D.5 Full Cross-Domain Strict Audit} Table~\ref{tab:cross_domain_full} reports the complete six-partition ledger underlying Figure~\ref{fig:cross_domain_aligned}. Default Credit remains nearly perfectly preserving across all four partitions, whereas both Heart Disease partitions exhibit measurable strict reversal.
\subsection{D.6 Controlled Strict-Audit Detail}
Table~\ref{tab:controlled_full} reports the main controlled sweep supporting Figure 1 of the main paper. Eligible-pair, strict-reversal, and pooled-tie columns are sums over the five replications, whereas the mismatch and strict-A2 columns are unweighted means of the five replication-level values used in Figure 1. The mean strict-A2 column is therefore not reconstructed by dividing the pooled reversal and tie counts by the pooled eligible count.

Near-perfect preservation at low mismatch gives way to a reversal-driven collapse at the observed transition. Pooled ties remain isolated, and the later non-monotone consistency values reinforce that the crossing is an evaluated transition location rather than a universal threshold.

\end{document}